%% file: neurips26_reasoning.tex
\documentclass{article}

\PassOptionsToPackage{numbers, compress, sort}{natbib}

\usepackage[preprint]{neurips_2026}

\usepackage[utf8]{inputenc} 
\usepackage[T1]{fontenc}    
\usepackage{hyperref}       
\usepackage{url}            
\usepackage{booktabs}       
\usepackage{amsfonts}       
\usepackage{nicefrac}       
\usepackage{microtype}      
\usepackage{xcolor}         
\usepackage{wrapfig}

\usepackage[most]{tcolorbox}
\usepackage{xcolor}
\usepackage{soul} 
\usepackage{enumitem} 

\definecolor{boxborder}{HTML}{555555}
\definecolor{boxbg}{HTML}{F7F7F7}
\definecolor{hlblue}{HTML}{CDE4F2}
\definecolor{hlorange}{HTML}{F4D3B6}
\definecolor{hlpurple}{HTML}{E2D4F0}
\definecolor{hlpink}{HTML}{F4C7D4}
\definecolor{hlyellow}{HTML}{F9E79F}
\definecolor{hlgreen}{HTML}{B9E4C0}
\definecolor{hlgrey}{HTML}{D3D3D3}

\newcommand{\hlblue}[1]{\sethlcolor{hlblue}\hl{#1}}
\newcommand{\hlorange}[1]{\sethlcolor{hlorange}\hl{#1}}
\newcommand{\hlpurple}[1]{\sethlcolor{hlpurple}\hl{#1}}

\newcommand{\hlyellow}[1]{\sethlcolor{hlyellow}\hl{#1}}
\newcommand{\hlgreen}[1]{\sethlcolor{hlgreen}\hl{#1}}

\newtcolorbox{promptbox}[1]{
  enhanced,
  colframe=boxborder,       
  colback=boxbg,            
  coltitle=white,           
  fonttitle=\bfseries\large,
  title=#1,                 
  boxrule=1pt,              
  arc=5pt,                  
  auto outer arc,
  left=0pt, right=0pt, top=0pt, bottom=0pt, 
  fontupper=\ttfamily\footnotesize,
  fontlower=\ttfamily\footnotesize,
  segmentation style={dashed, draw=boxborder, line width=0.5pt} 
}

\title{Robust Reasoning Benchmark}

\author{%
  Pavel Golikov$^{1,2}$, Evgenii Opryshko$^{1,2}$, Gennady Pekhimenko$^{1,2}$, Mark C. Jeffrey$^1$ \\
  $^1$University of Toronto, $^2$Vector Institute \\
}

\begin{document}

\input{floats.tex}

\maketitle

\input{0_abstract}

\input{1_intro}

\input{2_background}
\input{3_transformations}

\input{4_methodology}
\input{5_results}

\section{Limitations}
While our benchmark reveals significant structural fragility in current LLMs, our evaluation
primarily focuses on the domain of competitive mathematics. The extent to which these findings generalize to
other high-stakes reasoning tasks (e.g., legal, medical, coding) remains to be explored in future work.

\input{6_conclusion}

\section*{Acknowledgments}
We gratefully acknowledge Vector Institute and Digital Research Alliance of Canada for providing the computational
resources required for this study on their Killarney cluster.

\bibliographystyle{plainnat}
\bibliography{refs}


\appendix
\input{appendix}

\end{document}

%% file: floats.tex

\newcommand{\figAvgAccDrop}{
\begin{wrapfigure}{r}{0.45\textwidth}
  \centering
  \includegraphics[width=\linewidth]{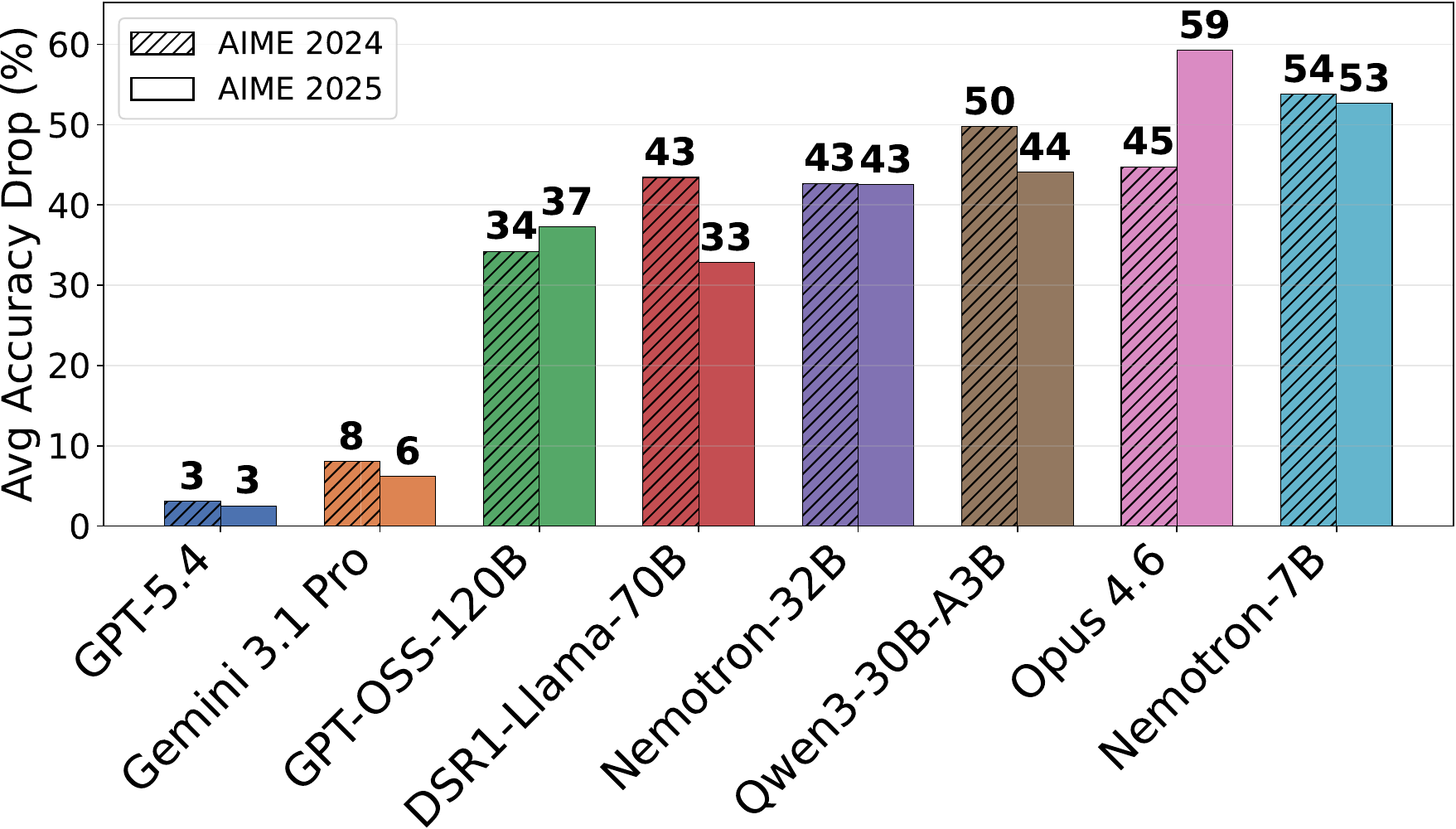}
  \caption{Average Accuracy Drop across all models and 13 structural perturbations for AIME 2024/2025 datasets.}
  \label{fig:avg_accuracy_drop}
\end{wrapfigure}
}

\newcommand{\figAccuracy}{
\begin{figure}[ht!]
  \centering
  \includegraphics[width=0.98\linewidth]{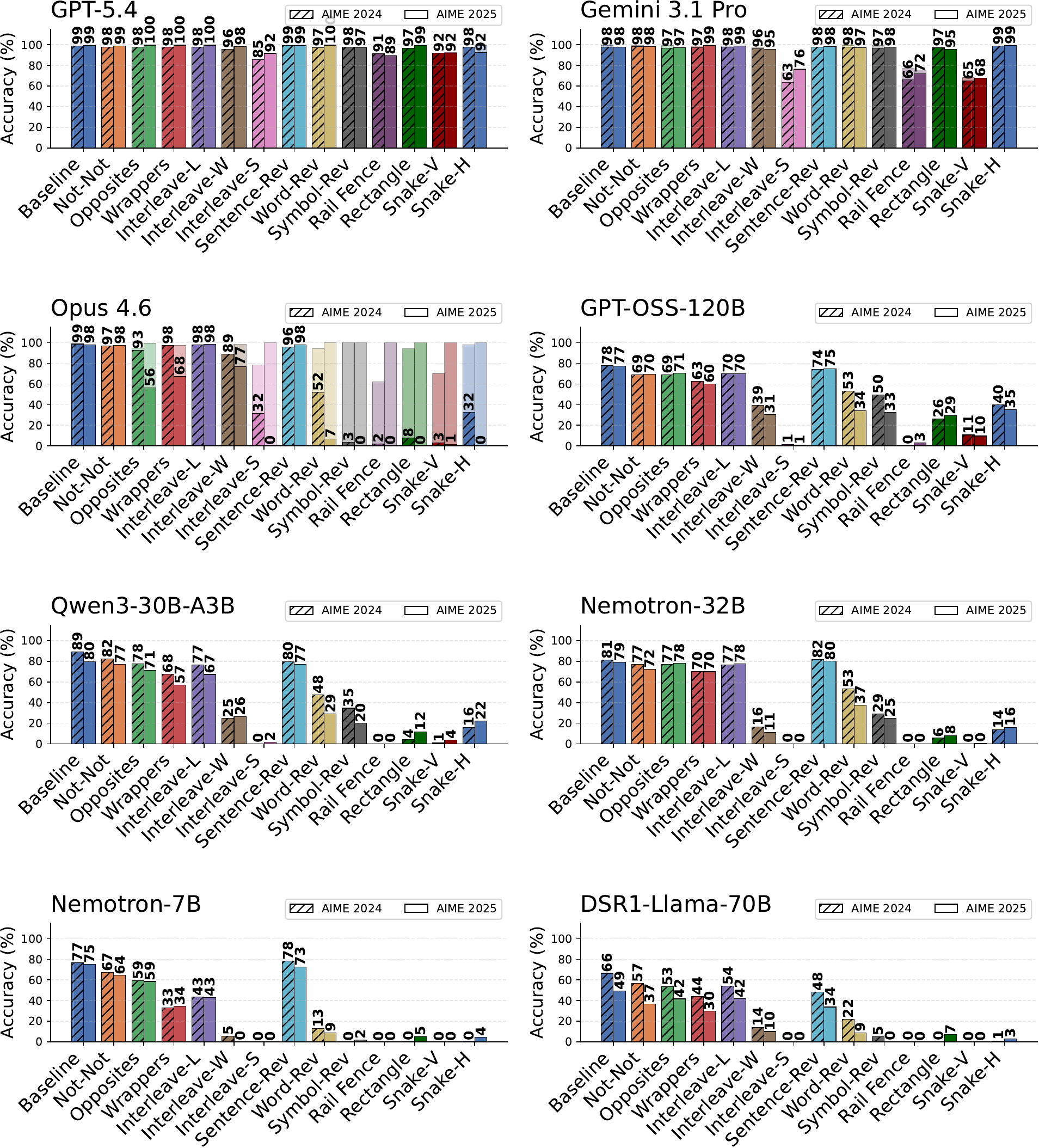}
  \caption{Full bars represent Achieved Accuracy on the AIME 2024/2025 benchmarks modified with 13 structural perturbations.
  Semi-transparent segments indicate model refusals or failed answers.} 
  \label{fig:accuracy_by_model}
\end{figure}
}

\newcommand{\figRadarCharts}{
\begin{figure*}[ht!]
  \centering
  \includegraphics[width=0.98\linewidth]{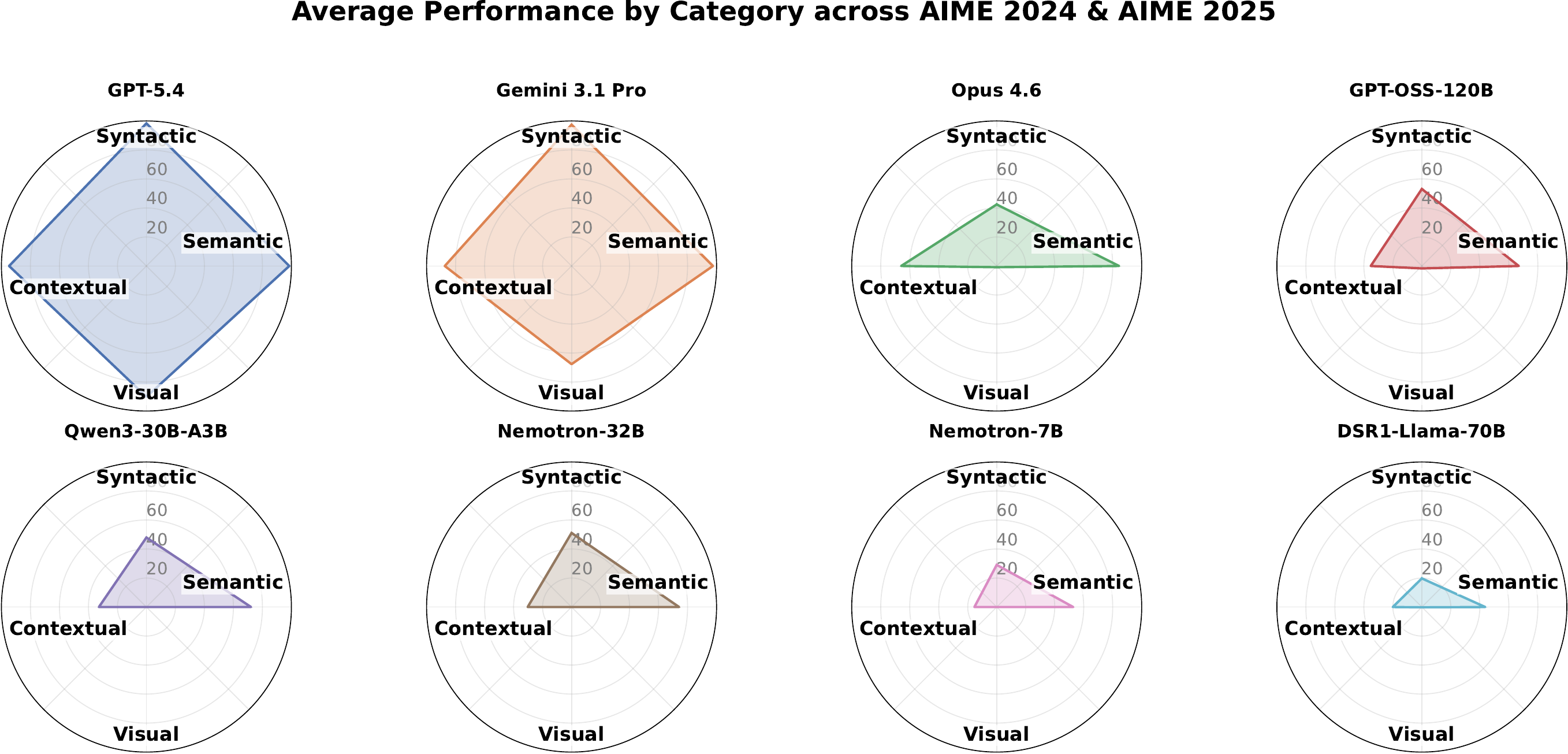}
  \caption{Vulnerability Profiling: Average accuracy on transformations from each of the 4 categories by model.}
  \label{fig:radar_charts}
\end{figure*}
}

\newcommand{\figOutputLength}{
\begin{figure}[ht!]
  \centering
  \includegraphics[width=0.98\linewidth]{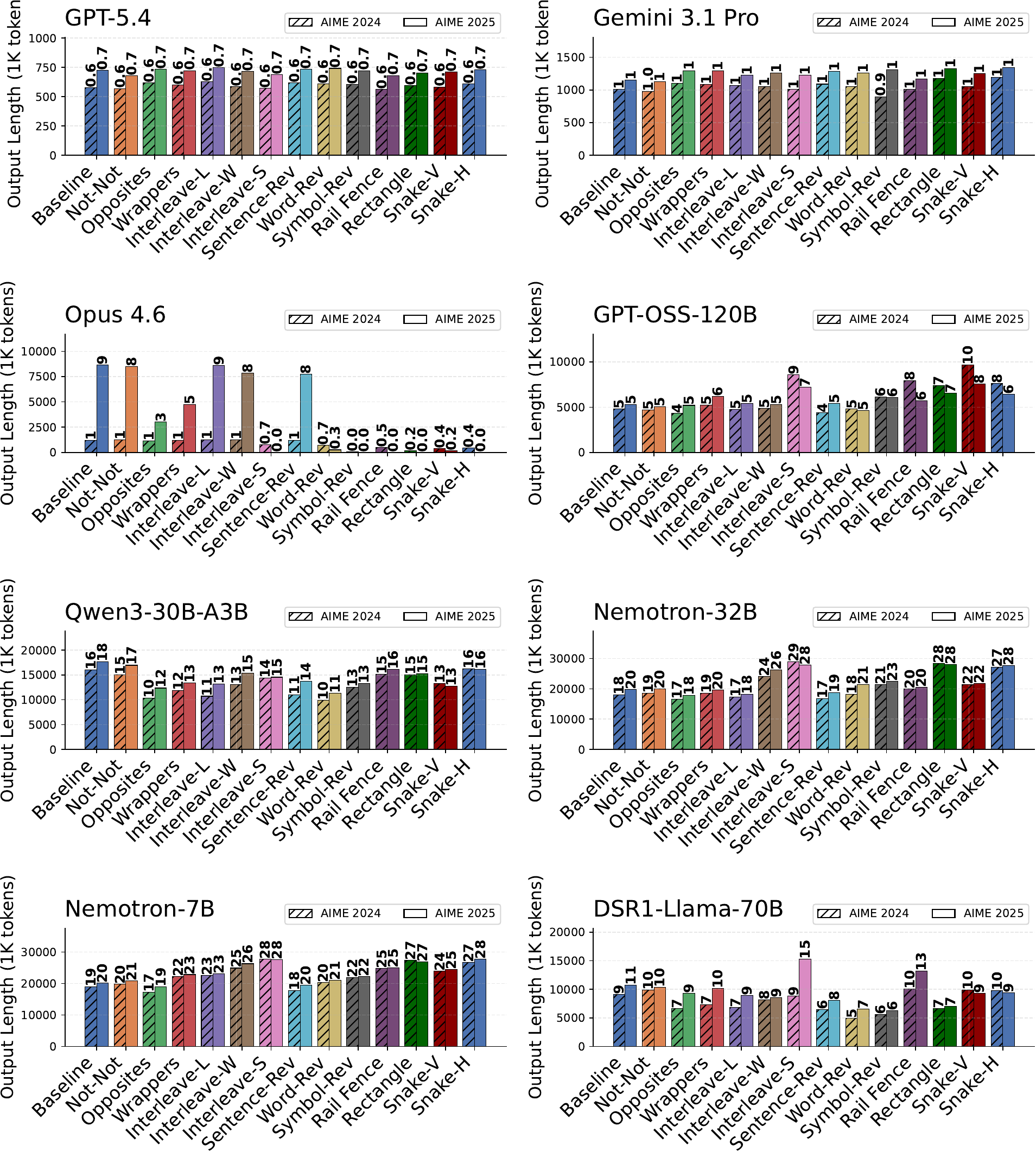}
  \caption{Reasoning Efficiency: Average output token length by task.
  On top of each bar is the average output length in thousands of tokens.}
  \label{fig:output_length}
\end{figure}
}

\newcommand{\figAttenHeatMap}{
\begin{figure}[ht!]
  \centering
  \includegraphics[width=0.98\linewidth]{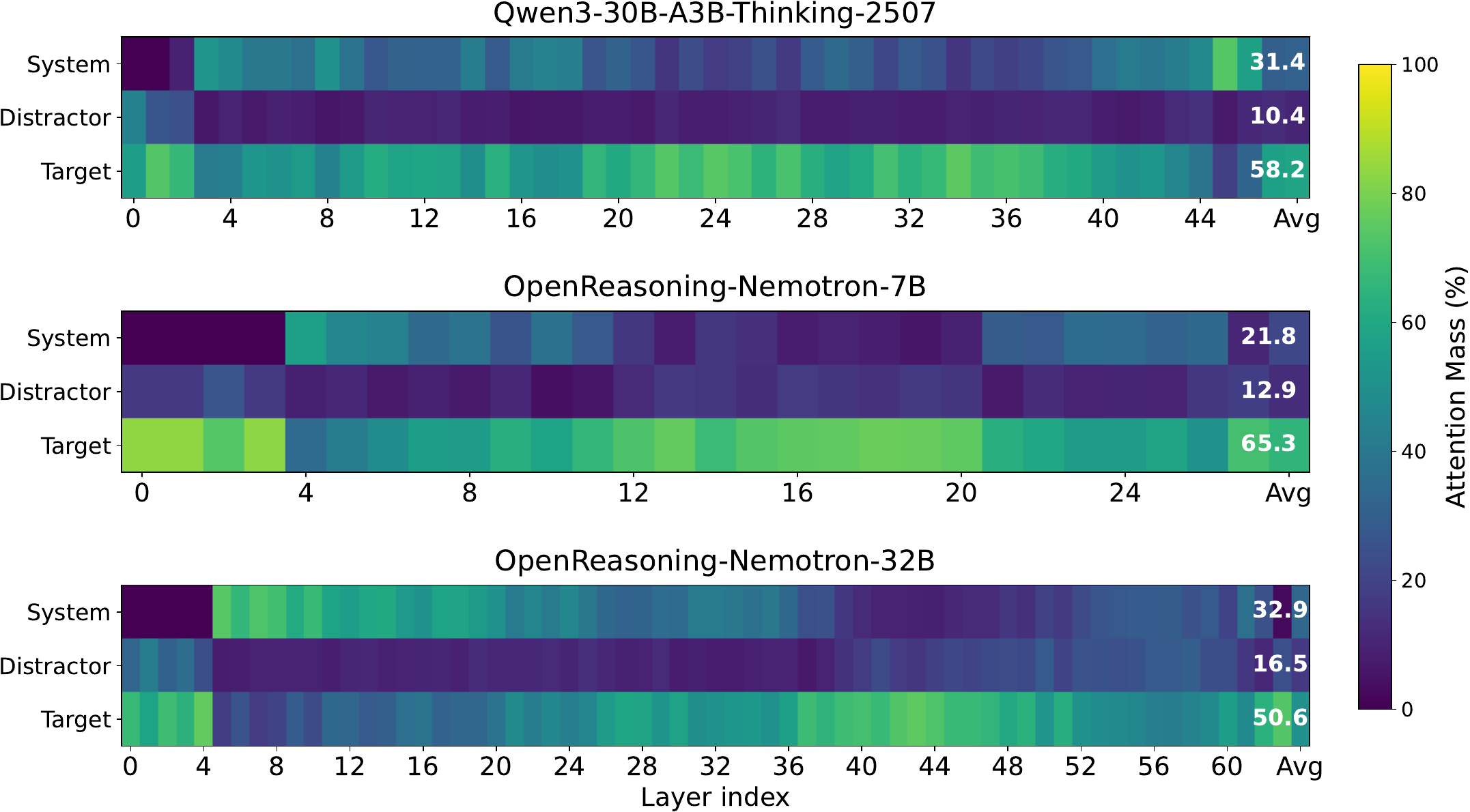}
  \caption{Attention Dilution Across Layers. Target attention scores averaged across samples, tokens, and heads,
  showing per-layer attention allocation over three context regions.
  The last x-label, "Avg", indicates the average across layers.}
  \label{fig:attention_dilution}
\end{figure}
}

%% file: 0_abstract.tex
\begin{abstract}
While Large Language Models (LLMs) achieve high performance on standard mathematical benchmarks,
their problem-solving abilities depend on the context and textual formatting.
We introduce the \emph{Robust Reasoning Benchmark (RRB)}, a pipeline of 13 deterministic textual perturbations
applied to AIME 2024 and AIME 2025. Evaluating 8 state-of-the-art models, we find that frontier models are largely
resilient, with the notable exception of Claude, which categorically refuses many transformed prompts. Open-weights
reasoning models exhibit a range of failure modes under structural noise (cognitive thrashing, tokenization breakdown,
and reasoning collapse), with up to 54\% average accuracy drops across perturbations and up to 100\% on some.
We further study one of these failure modes in isolation: attention dilution caused by the model's own
chain-of-thought. By tasking models with solving multiple independent mathematical problems sequentially within a
single context window, we identify \emph{Intra-Query Attention Dilution}. Open-weights models ranging from 7B to 120B
parameters exhibit accuracy decay on subsequent problems, suggesting that intermediate reasoning steps progressively
pollute standard dense attention mechanisms.
We argue that in order to achieve reliable reasoning, future architectures need to integrate explicit
contextual resets within models' own chain-of-thought, leading to open research questions regarding the optimal
granularity of reasoning tasks.
\end{abstract}

%% file: 1_intro.tex
\section{Introduction}
\label{sec:intro}

Large Language Models (LLMs) have achieved remarkable success on mathematical reasoning benchmarks, with
state-of-the-art models approaching saturation on datasets like AIME, GSM8K, and MATH~\citep{gsm8k, math, aime_2024}.
These results have fueled the perception that LLMs possess robust algorithmic reasoning capabilities: the ability to
decompose complex problems into logical intermediate steps. However, a growing body of literature~\citep{embers,
token_bias, alice} suggests that this performance may be brittle, relying instead on probabilistic pattern matching and
surface-level correlations. 

To distinguish between robust reasoning and fragile memorization, recent research has turned to adversarial evaluation.
Existing approaches generally fall into two categories: changing numeric values to test arithmetic
generalization~\citep{varbench, eval_robustness, numerical_sensitivity} or increasing environmental complexity by adding
irrelevant context or harder logic~\citep{gsm_ic, gsm_ir, problemathic}. While valuable, these methods often introduce
confounding variables. When a model fails after numerical perturbation, it is unclear whether the failure lies in the
reasoning logic or the arithmetic calculation. Similarly, when a model fails on a more complex problem, it is difficult
to distinguish between a lack of robustness and a hard limit on the model's inherent capacity. Furthermore, some
benchmarks rely on LLMs to generate the perturbations, introducing non-deterministic noise and potential validity errors
into the evaluation pipeline.

\begin{figure}[t]
    \centering
    \scalebox{0.9}{
    \begin{minipage}[c]{0.54\textwidth}
        \begin{tcolorbox}[
            enhanced,
            colframe=darkgray,
            colback=white!97!gray,
            coltitle=white,
            fonttitle=\bfseries\small,
            title={User Prompt Example},
            boxrule=1pt,
            arc=3pt,
            left=4pt, right=4pt, top=4pt, bottom=4pt,
            fontupper=\ttfamily\scriptsize\linespread{1.3}\selectfont,
            fontlower=\ttfamily\scriptsize\linespread{1.3}\selectfont
        ]
            Solve these completely unrelated math problems. For each problem put your final answer within
            \texttt{\textbackslash \textbackslash boxed\{\}}.\\
            Problem 1: Find the number of ways to place a  \dots\\
            Problem 2: Let $ABC$ be a triangle inscribed   \dots\\
            Problem 3: Let $p$ be the least prime number for\dots\\
        \end{tcolorbox}
    \end{minipage}
    }
    \hfill
    \begin{minipage}[c]{0.49\textwidth}
        \centering
        \includegraphics[width=\linewidth]{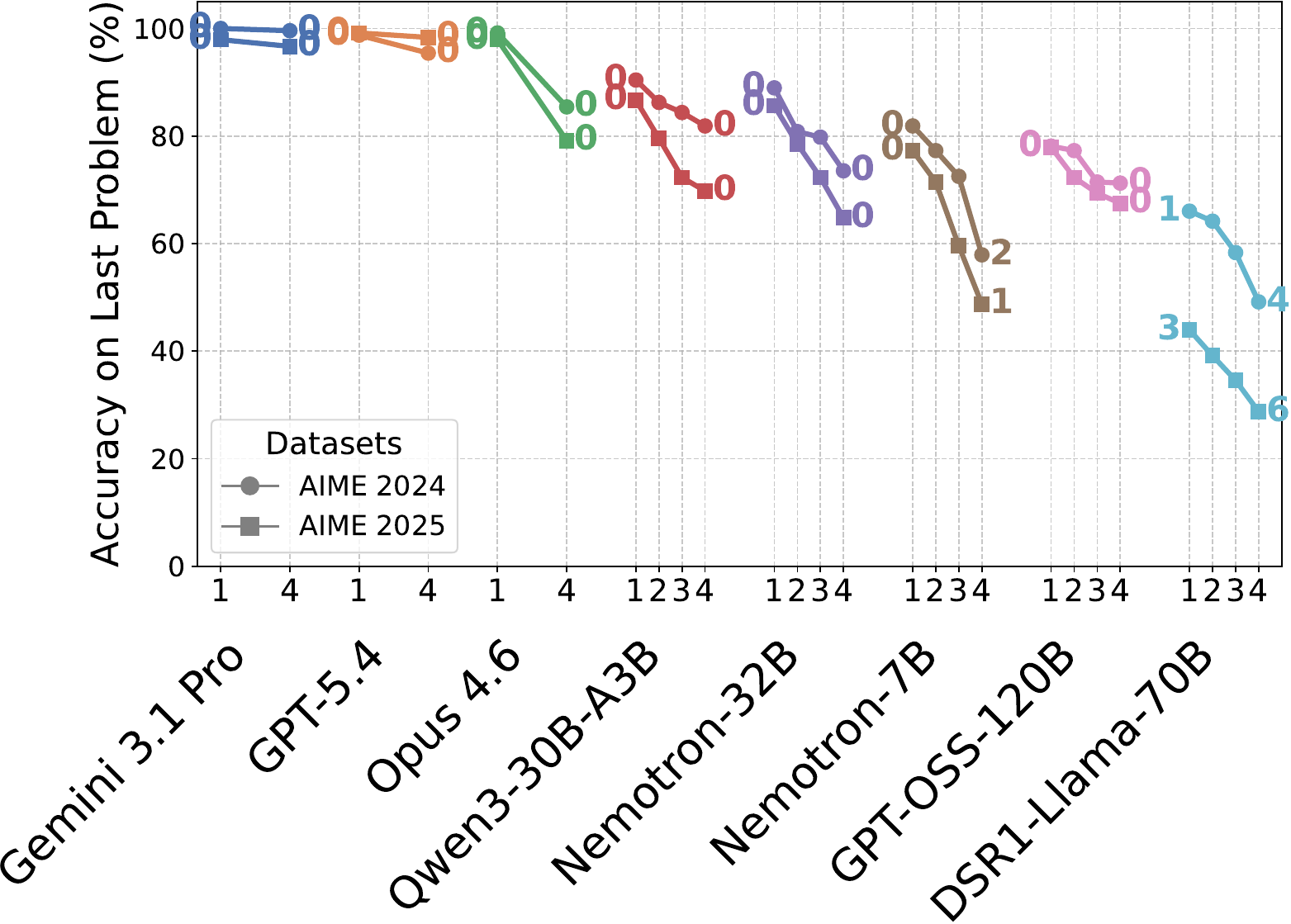}
    \end{minipage}
    \vspace{0.2cm}
    \caption{\textbf{Intra-Query Attention Dilution.} \textit{Left:} The sequential cognitive overload setup, where
    we prompt the models to solve multiple independent dataset questions within a single prompt.
    \textit{Right:} Results of the sequential cognitive overload experiment.
    X-axis markers indicate the total number of problems in the sequence. Values at the start and end of each
    line show the percentage of samples that reached the max token cutoff.
    Line plot markers show mathematical accuracy strictly on the \textit{final} problem of the sequence.
    While frontier APIs exhibit stronger resilience, all tested open-weights models
    suffer a degradation in reasoning accuracy.
    Results suggest that reasoning degradation is scale-invariant across open-weights models ranging
    from 7B to 120B parameters.
    This suggests that prior reasoning steps progressively degrade subsequent reasoning ability, highlighting an
    architectural need for explicit working memory isolation within the model's own chain-of-thought.}
    \label{fig:teaser}
\end{figure}

In this paper, we introduce the \textbf{Robust Reasoning Benchmark (RRB)}, an evaluation framework based on 13
deterministic structural transformations. Unlike prior work, we do not alter the mathematical values, the logic, or the
final answer. Specifically, we apply reversible algorithmic perturbations such as syntactic string reversals, visual 2D
grid encodings (e.g., Rail Fence, Horizontal Snake~\citep{rail_fence_cipher,boustrophedon}), and semantic wrappers.
We provide the description of each transformation in
context to isolate the ability to perform the reverse transformation from the ability to recognize the transformation in
the first place. Our transformations are trivial for humans to decode but structurally hostile to LLM tokenization and
attention. RRB exposes failure modes that vary by model class: frontier models are largely resilient, with the notable
exception of Claude, which categorically refuses many transformed prompts; open-weights reasoning models exhibit
cognitive thrashing, tokenization breakdown, and attention dilution from the model's own decoding chain-of-thought.
We further study the last of these failure modes in isolation. To remove transformation decoding as a confound,
we task models with
solving several unperturbed mathematical problems within a single query and measure accuracy only on the last problem.
This protocol identifies \textbf{Intra-Query Attention Dilution}. As visualized in Figure~\ref{fig:teaser}, accuracy on
the final problem degrades as the context window is polluted by the model's own prior reasoning steps.

Long-context attention drift is a recognized challenge in information retrieval~\citep{lost_in_the_middle,chroma_context_rot}.
Recent system-prompt leaks from proprietary agentic frameworks~\citep{antigravity_leaks} (see also
Appendix~\ref{sec:agent_context_leaks}) suggest the industry is likely attempting to mitigate this through the use of explicit
contextual resets.
In this work, we demonstrate that such pollution exists even within the span of a single query.
We argue that this represents a structural limitation of the standard dense attention mechanism, which
lacks reliable mechanisms for working memory isolation.
To support more reliable algorithmic reasoning, we argue that future architectures should use explicit contextual
resets within the model's own Chain-of-Thought. Whether implemented natively or by delegating steps to
sub-agents~\citep{subagents,recursive_lm}, enforcing these cognitive boundaries introduces an open research question
regarding the optimal \textit{granularity of reasoning} for a particular model and task, which we further discuss in
Appendix~\ref{sec:open_questions}.

In summary, our contributions are:
\begin{itemize}[nosep, leftmargin=*]
    \item The \emph{Robust Reasoning Benchmark (RRB)}, augmenting AIME 2024 and AIME 2025 with 13 deterministic
    textual transformations that perturb structure without altering meaning or difficulty.
    \item Identification of \emph{Intra-Query Attention Dilution} via a sequential cognitive overload protocol, with
    mechanistic attention analysis confirming that prior reasoning steps pollute the dense attention mechanism.
    \item A comprehensive evaluation of LLM reasoning reliability across 5 open-weights and 3 closed-source models.
\end{itemize}

%% file: 2_background.tex
\section{Background and Related Work}
\subsection{Reasoning Large Language Models}
Researchers have introduced several benchmarks to evaluate the performance of LLMs on mathematical reasoning tasks, such as
GSM8K, MATH, and AIME 2024~\citep{gsm8k,math,aime_2024}. Techniques like Chain-of-Thought (CoT) and Tree-of-Thought
(ToT)~\citep{cot,tot} have been the primary drivers of this success.
However, static benchmark accuracy can mask the fragility of the underlying thinking process. 

While Large Language Models (LLMs) excel at pattern matching, distinguishing this from genuine reasoning remains a
challenge. Following the classical definition by~\citet{newell_simon_1976}, reasoning involves the
manipulation of symbols to bridge premises and conclusions. In cognitive science, this aligns with algorithmic thinking
(a slow, deliberate, and multi-step process), as opposed to the rapid, intuitive processing often associated with
neural inference~\citep{kahneman_2011}.

A critical attribute of robust reasoning is invariance to surface-level perturbations. If a model truly understands
the concepts and how to manipulate them (like words, sentences, symbols, and math operations), it should be resilient
to both surface-level perturbations and deeper logical distractions.

\subsection{Prior Works}
\textbf{Surface-level and Informational Perturbations.}
A prominent line of prior work evaluates robustness by introducing irrelevant information or lexical shifts to test
whether a model is reasoning or merely pattern matching.
\citet{gsm_symbolic}, \citet{problemathic}, and others~\citep{gsm_ic, token_bias, gsm_ir}
demonstrate the fragility of LLM math reasoning by injecting irrelevant numeric variables or token biases.
Beyond irrelevant information, \citet{rupbench}, \citet{rommath}, and meaning-preserving rephrasing
studies~\citep{mathrobust_lv, putnam_gap, mscr, cultural_translation, ar_checker, mpa, agentforest} systematically
quantify how homophones, sentence rephrasings, and visual attacks degrade accuracy.
Unlike these methods, which can introduce semantic ambiguity or rely on LLM-generation, our perturbations
are strictly deterministic and operate purely at the textual level. By explicitly providing the decoding rule, we isolate pure
structural fragility without altering the semantic or logical constraints of the problem. 

\textbf{Dataset Contamination, Memorization, and Numerical Variations.}
Another approach alters values and variables to prove the model is not simply reciting training data. Works like
\citep{varbench, alice, gsm_plus, numerical_sensitivity, adv_mwp} dynamically alter numeric values or utilize abstract
syntax trees to minimally edit problems, while \citep{functional_benchmarks, knights_knaves} convert static problems into
code functions to enable automatic generation or measuring memorization limits. \citet{embers}
demonstrates that text probability heavily drives performance.
While changing numbers effectively tests memorization, it modifies the mathematical constraints themselves. Our
approach preserves the exact pristine mathematical state. Furthermore, unlike~\citet{embers}, which solely tests text decoding,
we force the model to sequentially decode \emph{and} reason, evaluating the combination of decoding structural noise
and downstream logic.

\textbf{Perturbations within Chain-of-Thought Trajectories.}
Recent work explores robustness by perturbing the intermediate steps of a model's reasoning trajectory.
\citet{cot_interventions} and \citet{fragile_thoughts} inject errors or interventions directly into
the CoT to highlight that internal reasoning processes remain susceptible to derailment.
Rather than perturbing the active reasoning steps, our work pollutes the initial problem presentation.

\textbf{Formal, Symbolic, and Difficulty-Based Evaluations.}
Several works modify existing benchmarks to evolve their inherent difficulty. \citet{algebraic_circuit}, \citet{asymob},
and others~\citep{illusion_thinking, ride, math_perturb,
ontology_guided} design controlled environments, perform confusing parameter substitutions, or rewrite questions into
more mathematically complex versions.  
These benchmarks intentionally increase the mathematical or algorithmic complexity of the task.
In contrast, our textual transformations maintain a constant level of difficulty.
Moreover, our method is entirely domain-agnostic and can be applied at scale to any existing LaTeX-formatted math
dataset without requiring complex problem rewrites.

%% file: 3_transformations.tex
\begin{figure*}[ht!]
\centering
\scalebox{0.8}{%
\begin{minipage}[t]{1.1\textwidth} 
\begin{minipage}[t]{0.48\textwidth}
    \begin{promptbox}{Semantic and Lexical Substitutions}
    {\bfseries\sffamily Original Problem}\\[2pt]
    Let \$p\$ be the least prime number for which there exists a positive integer \$n\$ such that \dots
    \tcbline
    {\bfseries\sffamily Not Not}\\[2pt]
    Let \$p\$ be the \hlblue{not not} least \hlblue{not not} prime \hlblue{not not} number for which there exists a \dots
    \tcbline
    {\bfseries\sffamily Opposites}\\[2pt]
    \hlorange{Forbid} \$p\$ be the \hlorange{most} \hlorange{antiprime} \dots \\
    \hlpurple{defyn\{let "most" mean "least", let "Forbid" mean "Let"\}.}\\ 
    \dots is \hlorange{indivisible} by \$p\^{}\{2\}\$.
    \tcbline
    {\bfseries\sffamily Wrappers}\\[2pt]
    Let \$\hlyellow{3(p)}\$ be the least prime \hlyellow{2(number)} for which there exists a \dots \\
    \hlgreen{defyn\{let "3(p)" mean "p", \dots let "7(p)" mean "p"\}.}\\ 
    \dots divisible by \$\hlyellow{7(p)}\^{}\{\hlyellow{9(2)}\}\$.
    \end{promptbox}
\end{minipage}%
\hfill
\begin{minipage}[t]{0.48\textwidth}
    \begin{promptbox}{Visual Encoding}
    {\bfseries\sffamily Rail Fence Cipher}\\[2pt]
    \hlorange{L}...\hlorange{\$}...\hlorange{b}...\hlorange{h}...\hlorange{e}...\hlorange{~}...\hlorange{m}...\hlorange{u}...\hlorange{r}...\hlorange{r}
    .\hlorange{e}.\hlorange{~}.\hlorange{p}.\hlorange{~}.\hlorange{e}.\hlorange{t}.\hlorange{e}.\hlorange{l}.\hlorange{a}.\hlorange{t}.\hlorange{p}.\hlorange{i}.\hlorange{e}.\hlorange{n}.\hlorange{m}.\hlorange{e}.\hlorange{~}.\hlorange{o}.
    ..\hlorange{t}...\hlorange{\$}...\hlorange{~}...\hlorange{~}...\hlorange{s}...\hlorange{r}...\hlorange{~}...\hlorange{b}...\hlorange{f}..
    \tcbline
    {\bfseries\sffamily Rectangle Perimeter}\\[2pt]
    \hlblue{Let \$p\$ be the least prime number for}\\
    ....................................~\\
    ....................................\hlblue{w}\\
    \hlblue{\$}...................................\hlblue{h}
    \tcbline
    {\bfseries\sffamily Snake Horizontal}\\[2pt]
    \hlgreen{$\rightarrow$ Let \$p\$ be the least prime number}\\
    \hlyellow{$\leftarrow$ vitisop a stsixe ereht hcihw rof }
    \tcbline
    {\bfseries\sffamily Snake Vertical}\\[2pt]
    \hlblue{L}\hlorange{e}\hlgreen{~}\hlpurple{t}\hlyellow{~}m~b~h~i~~~e~p~o~n~t~~~s~~~\$~~~i\\%
    \hlblue{e}\hlorange{b}\hlgreen{t}\hlpurple{s}\hlyellow{p}u~e~w~c~e~x~~~s~i~e~\$~u~t~n~\$~s\\
    \hlblue{t}\hlorange{~}\hlgreen{h}\hlpurple{a}\hlyellow{r}n~r~~~h~r~i~a~i~~~g~n~c~a~\^{~}~1~\\
    \hlblue{~}\hlorange{\$}\hlgreen{e}\hlpurple{e}\hlyellow{i}~~~~r~~~e~s~~~t~e~e~\$~h~h~\{~+~d\\
    \hlblue{\$}\hlorange{p}\hlgreen{~}\hlpurple{l}\hlyellow{m}e~f~o~t~h~t~s~i~v~r~~~~~t~4~\}~i~
    \end{promptbox}
\end{minipage}

\vspace{0.4cm}

\begin{minipage}[t]{0.48\textwidth}
    \begin{promptbox}{Contextual Overload}
    {\bfseries\sffamily Line-Level}\\[2pt]
    \hlblue{<Problem A> Let {\$}p{\$} be the least \dots}\\
    <Problem B> Let \$ABCD\$ be a tetr \dots\\
    \hlblue{<Problem A> ositive integer {\$}n{\$}~  \dots}
    \tcbline
    {\bfseries\sffamily Word-level}\\[2pt]
    \hlorange{Let} Let \hlorange{\$p\$} \$ABCD\$ \hlorange{be} be \hlorange{the} a \hlorange{least} tetrahedron \hlorange{prime}
    such \hlorange{number} that \dots
    \tcbline
    {\bfseries\sffamily Symbol-level}\\[2pt]
    \hlpurple{L}L\hlpurple{e}e\hlpurple{t}t\hlpurple{~}~\hlpurple{\$}\$\hlpurple{p}A\hlpurple{\$}B\hlpurple{~}C\hlpurple{b}D\hlpurple{e}\$\hlpurple{~}~\hlpurple{t}b\hlpurple{h}e\hlpurple{e}~\hlpurple{~}a\hlpurple{l}~\hlpurple{e}t\hlpurple{a}e\hlpurple{s}t
    
    \hlpurple{t}r\hlpurple{~}a\hlpurple{p}h\hlpurple{r}e\hlpurple{i}d\hlpurple{m}r\hlpurple{e}o\hlpurple{~}n\hlpurple{n} \hlpurple{u}s\hlpurple{m}u\hlpurple{b}c\hlpurple{e}h\hlpurple{r} \hlpurple{~}t\hlpurple{f}h\hlpurple{o}a\hlpurple{r}t\hlpurple{~} 
    

    \end{promptbox}
\end{minipage}%
\hfill
\begin{minipage}[t]{0.48\textwidth}
    \begin{promptbox}{Syntactic Distortions}
    {\bfseries\sffamily Sentence Reversal}\\[2pt]
    ~Find the least positive integer \$m\$ such that {\$}n{\^{}}\{4\}+1{\$} is divisible by {\$}p{\^{}}\{2\}{\$}. Let \$p\$
    be the least prime number \dots
    \tcbline
    {\bfseries\sffamily Word Reversal}\\[2pt]
    \$p\^{}\{2\}\$. by divisible is \$m\^{}\{4\}+1\$ that such \$m\$ integer positive least the Find
    \tcbline
    {\bfseries\sffamily Symbol Reversal}\\[2pt]
    teL \$p\$ eb eht tsael emirp rebmun rof hcihw ereht stsixe a evitisop regetni \$n\$ \dots
    \end{promptbox}
\end{minipage}
\end{minipage}}
\caption{Examples of the 13 structural transformations applied to a sample mathematical query. Target problems and lengthier text mappings have been truncated (\dots) for brevity.}
\label{fig:all_transformations}
\end{figure*}

\section{The Robust Reasoning Benchmark}
\label{sec:transformations}

To evaluate the robustness of mathematical reasoning, we design 13 input perturbations, grouped into four broad categories.
These transformations are strictly textual and deterministic; they are cognitively trivial for a human with access to pen, paper, and the decoding rules.
Figure~\ref{fig:all_transformations} shows examples.

\subsection{Semantic and Lexical Substitutions}
This category challenges the model's ability to maintain logical coherence when aliasing or redundant operators obscure
linguistic structures. 
\emph{Not Not:} We insert double negations (``not not'') prior to numerical values and
adjectives. Because the double negation is logically equivalent to the
original term, a robust reasoner should parse and discard these terms without altering its mathematical logic.
\emph{Opposites:} We remap semantic terms within the query to antonyms (e.g., ``short'' means ``long'', ``stop''
means ``goes''). 
\emph{Wrappers:} We encapsulate terms within arbitrary syntactic wrappers (e.g., replacing ``morning''
with ``3(morning)''). All wrappers are strictly identity functions.
For both \emph{Opposites} and \emph{Wrappers}, we explicitly define the specific localized substitutions inside
the user query, so decoding these problems requires simple key-value substitution.

\subsection{Contextual Overload}
\emph{Interleaved Contexts:} The text of two distinct math problems (Problem A and Problem B) is interwoven into a
single prompt. We explicitly instruct the model to only solve Problem A. We evaluate three granularities of interleaving.
\emph{Line-level:} We split the statements of two problems into line segments of at most 60 symbols.
We place each segment on a separate line, prefix it with a tag (e.g., \texttt{<Problem A>}), and strictly alternate segments for the two problems.
\emph{Word-level:} We strictly alternate the words of Problem A and Problem B one by one.
\emph{Symbol-level:} We alternatingly interweave the individual characters/symbols of the two problems.
If one problem statement is shorter than the other, we pad the remaining gaps by repeating the shorter problem from
its beginning.


\subsection{Syntactic Distortions}
In this category, the transformations split the input text into parts based on a specific delimiter (e.g., spaces or periods) and either reverse or rearrange the order of the resulting parts. We propose three variants:
\emph{Sentence Reversal:} We reverse the order of sentences (defined as sequences of symbols separated by periods)
in the user query.
\emph{Word Reversal:} We reverse the order of words (words defined as sequences of symbols separated by spaces) in the user query.
\emph{Symbol Reversal:} We reverse the symbols of every word (words defined as in Word Reversal above) in the user query.

\subsection{Visual and Spatial Encoding}
This category tests the model's character-level attention by mapping the 1D problem string onto a
2D visual grid. \texttt{GRID START} and \texttt{GRID END} markers bound all transformed inputs in this category. We evaluate four distinct spatial transformations.
\emph{Rail Fence Cipher:} We place the symbols of the encoded string in a zigzag pattern across multiple
rails (rows) and fill empty spatial gaps with dots (\texttt{.})~\citet{rail_fence_cipher}.
\emph{Rectangle Perimeter:} We map the user query onto the perimeter of a rectangle. The message follows the edges of the shape as a single continuous string in a clockwise manner, beginning at the top-left.
\emph{Snake Vertical:} We write the user query into a grid using a vertical `snake' (zigzag) pattern.
Starting from the top-left, the text flows down the first column, then up the second column, then down the third, and so on.
\emph{Snake Horizontal:} We write the user query into a grid using a horizontal `snake' (boustrophedon) pattern.

%% file: 4_methodology.tex
\section{Methodology}
\paragraph{Dataset Preprocessing and Sanitization.}
To ensure our evaluation strictly measures mathematical reasoning and avoid introducing low-level parsing or
tokenizer artifacts, we apply a preprocessing pipeline to AIME 2024~\citep{aime_2024} and AIME 2025~\citep{aime_2025} datasets.
String reversals can create executable escape sequences. For instance, if a
variable is enclosed in escaped parentheses, e.g., \verb|\(b\)|, character reversal yields \verb|\)\b(\|. During
string rendering, \verb|\b| may be interpreted as a backspace, thereby silently deleting the preceding
character. To eliminate information loss, we apply three preprocessing steps to all problem statements
prior to transformation. For completeness, we apply these steps to the baseline as well.
\emph{LaTeX Comment Removal:} We strip all inline LaTeX comments.
\emph{Newline Flattening:} We replace all newline characters with a semicolon followed by a space (\verb|;|). This
preserves the logical segmentation of the text while preventing spatial layout anomalies that can confuse 
the model's spatial attention during transformation reversal.
\emph{Escape Sequence Neutralization:} We systematically insert a space between backslashes and specific
characters (\verb|n, t, b, r, a, f|) to neutralize the accidental formation of control characters during string
reversal. These sanitization steps are semantically invariant; they do not meaningfully alter the mathematical problem and are
trivial for a human reader to parse and ignore. We apply these steps to all transformations, including the baseline. 

\paragraph{Transformation Design Principles.}
To ensure rigor, fairness, and consistency, every adversarial transformation adheres to the following guidelines:
\emph{Information-Theoretic Invariance:} The transformation cannot add extraneous mathematical constraints, remove
necessary premises, or introduce lexical ambiguities that would change the difficulty of the problem.
\emph{Algorithmic Reversibility:} The transformation must be deterministic and programmatically invertible.
We verify this by using Python reversal scripts for each transformation rule, confirming that the
original problem statement can be reconstructed without heuristics.
\emph{Cognitive Tractability:} The transform must be simple enough that a human, equipped only with pen and
paper, can decode the text exactly.
\emph{Explicit Decoding Protocol:} We provide the exact transformation rule in plain English to the model in the user
query. We explicitly instruct the model to execute a two-stage trajectory: first, programmatically decode the input to
reconstruct the problem statement; second, execute mathematical reasoning to solve it.
\emph{Grid Padding and Alignment:} For all visual and spatial encodings, we ensure that character spacing is
strictly preserved by substituting blank spaces with explicit dots (\texttt{.}) or padding the grid dimensions
perfectly. This ensures that when the LLM's tokenizer parses the grid, the geometric alignment (e.g., columns in the
vertical snake, or edges in the rectangle) remains structurally intact. 

We supply all models with a standard chain-of-thought system prompt: "You are a helpful math assistant. Please reason
step by step, and put your final answer within \verb|\\boxed{}|."
To disentangle perturbation decoding from reasoning failures, we perform a sequential cognitive overload experiment, where
we ask the model to solve a set of problems within a single query and record the accuracy only on the last problem.
For sequential cognitive overload, we additionally
prompt the model in user query with "Solve these completely unrelated math problems. For each problem put your final
answer within \verb|\\boxed{}|".

\paragraph{Evaluation Metrics.}
We evaluate the robustness of each model using two primary metrics.
\emph{Problem-Solving Accuracy:} The percentage of final mathematical answers evaluated as strictly correct
(extracted via the \verb|\\boxed{}| format using the \texttt{math\_verify}~\citep{math_verify} package). We count outputs that reach
the maximum token limit as failures.
For sequential cognitive overload, we search through all \verb|\\boxed{}| blocks and count the sample as
correct if any of them match the ground truth of the target problem, in case the model outputs answers out of order.
\emph{Reasoning Efficiency:} The token consumption required by the model to reach a final solution,
serving as a proxy for the computational friction introduced by the adversarial transformations. We defer discussion of
Reasoning Efficiency to Appendix~\ref{sec:reasoning_efficiency}.

\paragraph{Mechanistic Evaluation of Attention Dilution.}
To mechanistically evaluate intra-query attention dilution, we trace the internal attention allocation during the
autoregressive generation of the solutions for select AIME 2025 problems and for select models. We use samples with
three distractors and the fourth target problem.
We partition the full sequence of generated and prompt tokens,
$X$ of length $N$, into three disjoint index sets based on structural boundaries:
\emph{Instruction/Sink ($I_{\mathrm{sys}}$):} Tokens belonging to the system prompt and general format
instructions. This region natively absorbs the \textit{attention sink} phenomenon common in autoregressive transformers.
\emph{Distractor ($I_{\mathrm{dist}}$):} Tokens comprising the irrelevant prior mathematical problems and
their generated chain-of-thought trajectories.
\emph{Target ($I_{\mathrm{tgt}}$):} Tokens comprising the final target problem and its actively generated
chain-of-thought. We extract the token indices for each of these regions by using the tokenizer with offset mapping.
Let $A^{(l,h)} \in \mathbb{R}^{N \times N}$ denote the causal attention probability matrix for layer $l$ and head $h$.
For each token $t \in I_{\mathrm{tgt}}$ actively generated while solving the target problem,
the probability mass allocated to a specific context
region $R \in \{I_{\mathrm{sys}}, I_{\mathrm{dist}}, I_{\mathrm{tgt}}\}$ is:
\begin{equation}
    \mu_R^{(l,h)}(t) = \sum_{j \in R} A^{(l,h)}_{t,j}
\end{equation}
To quantify the macro-level architectural distraction at a given layer $l$, we average this probability mass over all
 $H$ attention heads and all target tokens $T = |I_{\mathrm{tgt}}|$:
\begin{equation}
    \bar{\mu}_R^{(l)} = \frac{1}{H \cdot T} \sum_{h=1}^H \sum_{t \in I_{\mathrm{tgt}}} \mu_R^{(l,h)}(t)
\end{equation}

\paragraph{Models and Experimental Scope.}
Our evaluation encompasses a suite of eight models, comprising five open-weights models and
three proprietary APIs. The open-weights models are: \texttt{Qwen/Qwen3-30B-A3B-Thinking-2507}~\citep{qwen3_thinking},
\texttt{nvidia/OpenReasoning-Nemotron-7B}~\citep{nemotron_7b},
\texttt{nvidia/OpenReasoning-Nemotron-32B}~\citep{nemotron_32b},
\texttt{DeepSeek-R1-Distill-70B}~\citep{deepseek_r1} (DSR1-Llama-70B), and
\texttt{openai/gpt-oss-120b}~\citep{gpt_oss_120b}.
The closed-source models are Gemini 3.1 Pro~\citep{gemini-3.1-pro}, GPT-5.4~\citep{gpt-5.4}, and Claude 4.6
Opus~\citep{claude-4.6-opus}.
We conduct all experiments on the AIME 2024~\citep{aime_2024} and AIME 2025~\citep{aime_2025} datasets.
We sample 8 zero-shot trajectories per problem for the proprietary models, and 16 zero-shot trajectories for open-weights
models. We use GPT-5.4 with the \texttt{xhigh} reasoning effort setting.
We use Claude 4.6 Opus with \texttt{max} reasoning effort and \texttt{adaptive} thinking enabled.
For experiments with perturbations, we set the temperature for all models to 0.7, \texttt{top\_p} to 1.0, and \texttt{max\_tokens} to 32K.
For sequential cognitive overload, we set the temperature to 0.6, \texttt{top\_p} to 0.95, and \texttt{max\_tokens} to 128K.
When evaluating sequential cognitive overload on proprietary models, due to high computational cost, we sampled only the
baseline and 4 sequential problems.
For attention extraction, we used open-weights models excluding \texttt{DeepSeek-R1-Distill-70B} and \texttt{openai/gpt-oss-120b} due to hardware limitations
for running attention extraction on large models.
We conducted all experiments on an internal cluster using up to 4 $\times$ NVIDIA H100~\citep{h100} GPUs per node,
with up to 128GB of CPU RAM. The cluster ran Ubuntu 24.04.2 LTS~\citep{ubuntu-24}.
We tested all open-weight models using vllm~\citep{vllm} version 0.13.0 and PyTorch 2.9.0.~\citep{pytorch}

%% file: 5_results.tex
\section{Results and Discussion}
\label{sec:results}
We evaluate mathematical reasoning robustness across our 13 input transformations and the sequential cognitive overload protocol.
Our results expose fragility in modern mathematical reasoning models, particularly within open-weights.

\figAccuracy

\subsection{Robustness Against Perturbations: Frontier vs. Open-Weights Models.}
Our macro-level evaluation reveals a stark divergence in architectural resilience towards textual perturbations,
visualized in the Average Accuracy Drop (Figure~\ref{fig:avg_accuracy_drop}) and Achieved Accuracy
(Figure~\ref{fig:accuracy_by_model}). State-of-the-art proprietary models exhibit remarkable robustness to 
structural noise; GPT-5.4 and Gemini 3.1 Pro suffer average performance degradations of only 3\% and 8\%,
respectively. In contrast, the open-weights ecosystem experiences substantial reasoning degradation.
Despite demonstrating strong baseline accuracy, models such as Nemotron-7B and
Qwen3-30B-A3B-Thinking-2507 suffer average accuracy drops of 53\% and 47\%, respectively.

\figAvgAccDrop
This ``robustness gap'' indicates that while current open-weights architectures and RL paradigms can memorize and
execute standard mathematical heuristics, their reasoning pathways are overfit to standard textual formatting and
sequential presentation.
All models handle semantic substitutions
relatively well across the board. However, open-weights models universally degrade on \emph{Syntactic Distortions}
and \emph{Visual Encodings}. Transformations that break words down into isolated characters, such as \textit{Interleave
(Symbol)} and \textit{Rail Fence Cipher}, strongly affect Byte-Pair Encoding (BPE) boundaries. Consequently, models
like Qwen3 and DSR1 drop to 0\% accuracy on these tasks. This shows that their general reasoning is bottlenecked
by subword semantic priors; they cannot map isolated character tokens back into a coherent internal causal graph. 
For example, on AIME 2024, when presented with the \textit{Snake (Horizontal)} task, Gemini 3.1 Pro
scores 99\%, and Qwen3 scores 16\%. However, when the exact same text is mapped vertically (\textit{Snake (Vertical)}),
Gemini drops to 65\%, and Qwen3 drops to near 1\%. This suggests that 
even when explicitly prompted to perform simple spatial grid algorithmic tasks, LLMs appear constrained by
their left-to-right, 1D pre-training biases.

\textbf{Overrefusal Issues.}
While closed-source models generally excel, Claude 4.6 Opus presents a striking anomaly, suffering a 52\% average
accuracy drop. A review of the data reveals that this drop is not due to reasoning
failure, but rather a disproportionate rate of model refusals (indicated by the semi-transparent upper segments of the bar
charts in Figure~\ref{fig:accuracy_by_model}). Claude categorically refuses to process the vast majority of the
prompts involving symbol manipulation. We hypothesize that this is an
artifact of its safety filters, which aggressively flag abstract symbol manipulation as prompt injection or jailbreak
attempts. This highlights an unintended consequence of current alignment strategies that penalize abstract,
character-level reasoning by misclassifying textually complex inputs as adversarial attacks.

\figAttenHeatMap
\subsection{Intra-Query Context Management}
\label{sec:architectural_implications}
Prior work establishes that long-horizon, multi-turn agentic workflows suffer from attention drift,
necessitating explicit state management to maintain coherence.
Complex mathematical problems, much like real-world tasks, require multi-step, sequential reasoning. As models become
more capable, they are used for ever more complex multi-step tasks.
Data in Figure~\ref{fig:teaser} suggest that this phenomenon is not
restricted to macro-level agentic interactions; it appears as a structural bottleneck that can degrade
reasoning \textit{within a single query}.
All evaluated open-weights models ranging from 7B parameters to 120B parameters experience this degradation in performance.
This suggests that intra-query attention dilution is a structural limitation of the standard
dense attention mechanism, which forces subsequent reasoning steps to unconditionally attend to prior ones.

Figure~\ref{fig:attention_dilution} illustrates how attention mass is distributed across the System, Distractor, and
Target regions during the generation of the target solution. The layer-wise breakdown highlights a transition:
while initial layers are focused on the Target problem, this focus quickly deteriorates.
Deeper layers exhibit classic "attention sink"
behavior, offloading significant amounts of probability mass (up to 32.9\% on average) to the initial System prompt. More
critically for reasoning robustness, the middle and late layers fail to isolate the target logic. Instead, they
consistently bleed attention into the Distractor region, maintaining a persistent 10\% to 16.5\% allocation to irrelevant
prior problems. This suggests that past reasoning trajectories act as a cognitive load within the dense attention mechanism. 


Recent community leaks~\citep{antigravity_leaks} and artifacts observed during the development of this project
(Appendix~\ref{sec:agent_context_leaks}) reveal that proprietary agent frameworks, such as Google's
AntiGravity~\citep{antigravity}, likely utilize explicit scaffolding like the \texttt{<task\_boundary\_tool>}.
This strongly suggests the industry is aware of long-context attention
drift and likely attempts to mitigate it by enforcing context management for long-horizon, multi-turn tasks.
Building on our empirical evidence of intra-query attention dilution, we argue that these same principles should
apply at the micro-level.
To support more reliable algorithmic reasoning, future architectures could benefit from mechanisms for
\textit{micro-context isolation}, allowing the model to explicitly summarize state and flush the noisy ``scratchpad'' of
previous intermediate steps, effectively resetting its attention focus within a single Chain-of-Thought and allowing it 
to focus on the substep at hand. This raises open research questions about optimal granularity of reasoning tasks,
which we discuss in Appendix~\ref{sec:open_questions}.

%% file: 6_conclusion.tex
\section{Conclusion}
In this work, we demonstrate that current LLMs suffer from mathematical reasoning fragility, revealing that their
reasoning prowess is often overfit to standard textual presentation rather than abstract logical manipulation.
By decoupling deterministic mechanical deciphering from downstream mathematical reasoning, we identify
intra-query attention dilution.
We empirically show that this reasoning degradation is present in open-weights models from 7B to 120B parameters,
demonstrating that intermediate reasoning steps pollute the context window.
Our findings, corroborated by artifacts leaked from proprietary agentic frameworks, suggest that standard dense attention
mechanisms struggle to support robust algorithmic reasoning because they fail to isolate working memory between
sequential reasoning steps \textit{within a single chain-of-thought}.
We argue that achieving reliable multi-step deduction requires the integration of native contextual resets
and task-boundary compartmentalization to flush intermediate states and preserve logical coherence.

%% file: appendix.tex
\section{Appendix}
\subsection{A: Methodology Details}
\label{sec:appendix_methodology_details}
When the model receives a transformed user query, we also provide instructions describing the transformation
that we applied to the original user query. Below are the transformation descriptions that we supply to the model in plain English.
The 'Not Not' transformation did not require any special decoding instructions, and we evaluate it using the standard
chain-of-thought prompt.
\begin{enumerate}
    \item Word Reversal:     The order of words (words are defined as sequences of symbols separated by spaces) in the user query has been reversed.
    \item Sentence Reversal: The order of sentences in the user query has been reversed. Sentences are defined as sequences of symbols separated by periods.
    \item Interleaved Context Word: User query will consist of two problems, A and B, whose statements are interleaved word by word. First word belongs to problem A, second word belongs to problem B, third word belongs to problem A, and so on. You need to solve only problem A. Words are defined as sequences of symbols separated by spaces. If one problem statement is shorter than the other, the empty spaces resulting from the shorter problem statement will be filled with the shorter problem statement repeated from the beginning.
    \item Interleaved Context Line: User query will consist of two problems, A and B, whose statements are split into line segments at most 60 symbols long. Each segment is followed by a space and a problem tag (e.g. problem A or B). The segments are interleaved. You need to solve only problem A. If one problem statement is shorter than the other, the empty lines resulting from the shorter problem statement will be filled with the shorter problem statement repeated from the beginning.
    \item Interleaved Context Symbol: User query will consist of two problems, A and B, whose statements are interleaved symbol by symbol (including punctuation and spaces). First symbol belongs to problem A, second symbol belongs to problem B, third symbol belongs to problem A, and so on. You need to solve only problem A. If one problem statement is shorter than the other, the shorter problem statement will be repeated from the beginning to fill the remaining space.
    \item Symbol Reversal:  Every word (words are defined as sequences of symbols separated by spaces) in user query has its symbols in reverse order.
    \item Opposites: There will be terms remapped in the user query. The remappings are defined inside 'defyn{}' block in the middle of user query.
    \item Wrappers:  There will be terms remapped in the user query. The remappings are defined inside 'defyn{}' block in the middle of user query.
    \item Rail Fence: The user query is encoded using the Rail Fence Cipher. The input is provided as a visual grid where the symbols (including spaces) of the encoded message string (message string does NOT contain any newline characters) are placed in a zigzag pattern across multiple rails (rows), and empty spaces are filled with dots (.). To decode, read the characters in zigzag order: Down-and-Right diagonally until you hit bottom rail, then Up-and-Right diagonally until you hit top rail, then Down-and-Right again etc... Rows are given on separate lines and all have equal lengths.
    \item Rectangle Perimeter: The user query is mapped onto the perimeter of a rectangle. The message is written as a single continuous string following the edges of the shape in a clockwise manner, beginning at the top-left. The TRANSFORMED INPUT is provided as a visual text block representing this rectangle with GRID START and GRID END markers. The center of the shape is filled with dots.
    \item Snake Vertical: The user query is written into a grid using a vertical 'snake' (zigzag) pattern. Starting from the top-left, the text is written down the first column, then up the second column, then down the third, and so on. The TRANSFORMED INPUT is provided as a visual grid with GRID START and GRID END markers.
    \item Snake Horizontal: The user query is written into a grid using a horizontal 'snake' (zigzag) pattern. Starting from the top-left, the text is written across the first row, then left across the second row, then right across the third, and so on. The TRANSFORMED INPUT is provided as a visual grid with GRID START and GRID END markers.
\end{enumerate}

\textbf{Disclosure:} This research utilized the AntiGravity agent framework to
programmatically develop the evaluation codebase and assist in experimental data processing.
The observational artifacts presented in Appendix~\ref{sec:agent_context_leaks} were captured during the development of
this codebase.

\figOutputLength

\subsection{Prompts}
\label{sec:prompts}
Baseline system prompt: `You are a helpful math assistant. Please reason step by step, and put your final answer within \verb|\\boxed{}|.'\\
Solution Protocol Prompt, inserted before the query:\\

YOUR PROTOCOL:\\
1. Read the "TRANSFORMATION RULE" provided by the user and reverse the transformation on the "TRANSFORMED INPUT" to obtain the original problem statement.\\
2. Once you have the original problem statement, proceed to solve the math problem.\\
3. Put your final answer within \verb|\\boxed{}|.

\subsection{Cognitive Thrashing}
\label{sec:reasoning_efficiency}
Figure~\ref{fig:output_length} shows the average output token length for each transformation.
Analysis of the figure reveals a pathological failure mode unique to
modern open-weights ``Thinking'' models (e.g., Qwen3, Nemotron). Rather than failing cleanly when presented with
unrecognizable structural logic, these models enter massive, unproductive loops. For example, when attempting the \textit{Rectangle Perimeter} task, Nemotron-7B outputs an average of 27K tokens, only to achieve 0\% accuracy. Similarly, Qwen generates 13K tokens on the \textit{Snake (Vertical)} task with 1\% accuracy. This indicates that the stopping criteria and logical progression heuristics are highly brittle.
When structural noise breaks the expected syntactic progression of a math problem, the model's internal
confidence mechanisms fail, resulting in hallucination loops rather than structured problem-solving.

\subsection{Limitations and Open Questions: The Granularity of Atomic Reasoning}
\label{sec:open_questions}

If the mitigation of intra-query attention dilution requires explicitly breaking complex reasoning into isolated
sub-tasks with contextual resets, it immediately raises a foundational open question for future research: \textit{What
is the optimal granularity of a sub-task?}
Currently, task decomposition in LLMs relies on arbitrary, heuristic-driven prompting techniques. While asking a
model to "think step by step" superficially mimics human deduction, models are neither trained for, nor equipped with,
the architectural mechanisms necessary for explicit subtask memory isolation and context management. This limitation
stems from their training corpora: models learn from step-by-step reasoning trajectories authored by humans, for
humans. In these demonstrations, the "under the hood" cognitive mechanics of task decomposition are entirely
abstracted away, as humans manage their own working memory boundaries implicitly. Forced to process sequential
logic in a continuous, unpartitioned state, models inevitably accumulate interference. Consequently, our findings
suggest that there exists a theoretical "atomic unit" of reasoning: the maximum amount of computation a model can
perform in a single context state before structural noise substantially degrades its algorithmic logic.

This introduces a critical trade-off. If a task is not decomposed enough, the model falls victim to the attention dilution
and cognitive thrashing demonstrated in our benchmark. Conversely, if a task is decomposed too finely, the system will
incur massive computational overhead from constantly explicitly summarizing state, calling boundary tools, and flushing
context, which may disrupt the continuous latent representations necessary for deep reasoning.

Furthermore, it remains unknown how this atomic reasoning capacity scales. It is entirely possible that the
\textit{optimal chunk size} of a sub-task scales conditionally with parameter count or network depth. We argue that formally quantifying the maximum safe
"cognitive load" of a single reasoning sub-task, together with designing architectures that can dynamically determine their
own optimal sub-task boundaries, represents an important frontier for the next generation of reasoning models.

While our work identifies structural fragility and attention dilution in the mathematical domain,
further study is required to determine if these effects generalize to non-numeric reasoning tasks and to
formally quantify the maximum safe 'cognitive load' of a single transformer forward pass.

\subsection{Agent Leaks}
\label{sec:agent_context_leaks}
We observed the following AntiGravity leaks during work on this project.
\begin{figure}[ht!]
    \centering
    \includegraphics[width=0.95\linewidth]{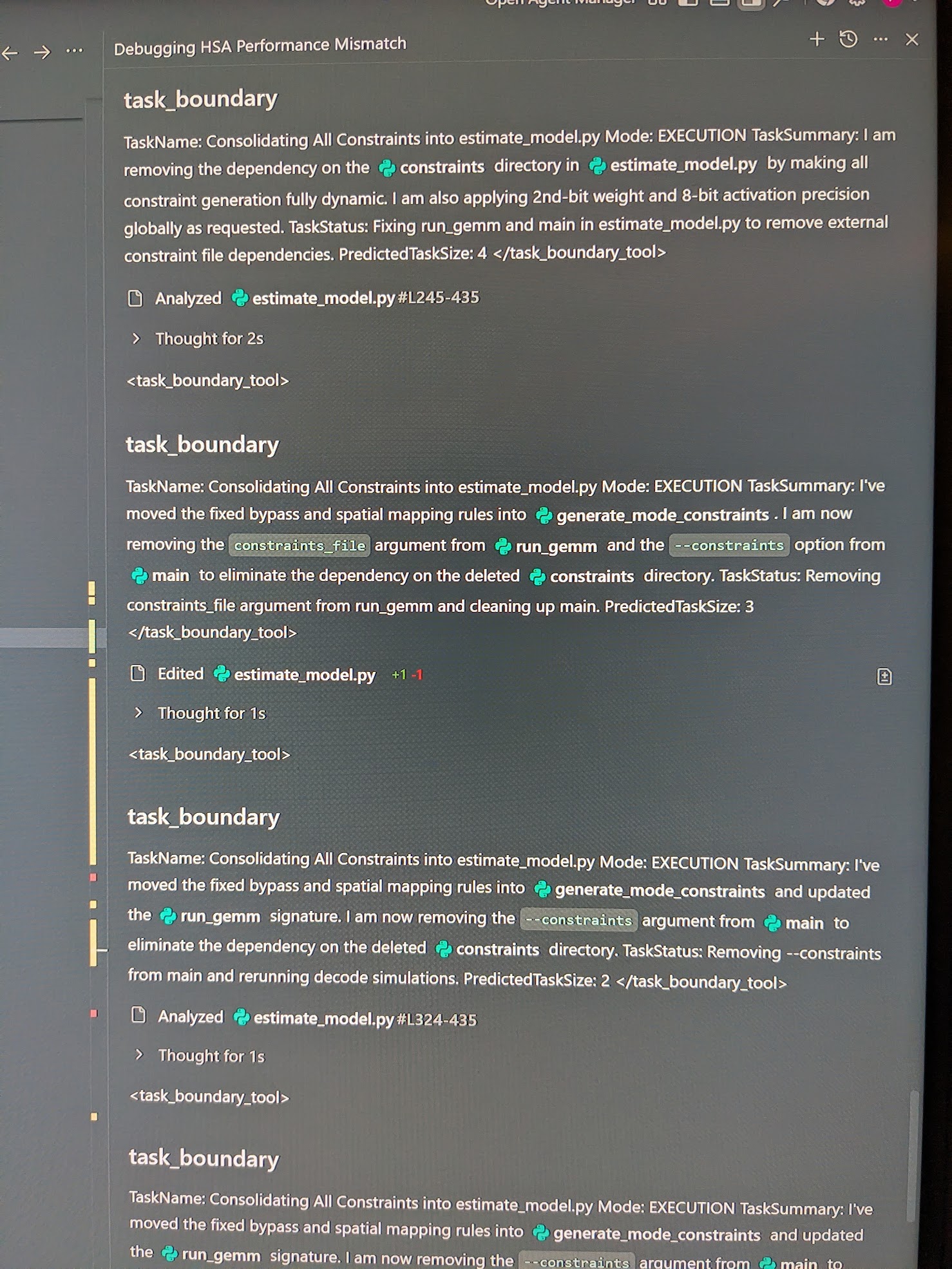}
    \caption{Agent Context Leaks (1/3): Example of the \texttt{task\_boundary} mechanism observed in AntiGravity.}
    \label{fig:agent_leaks}
\end{figure}

\begin{figure}[ht!]
    \centering
    \includegraphics[width=0.95\linewidth]{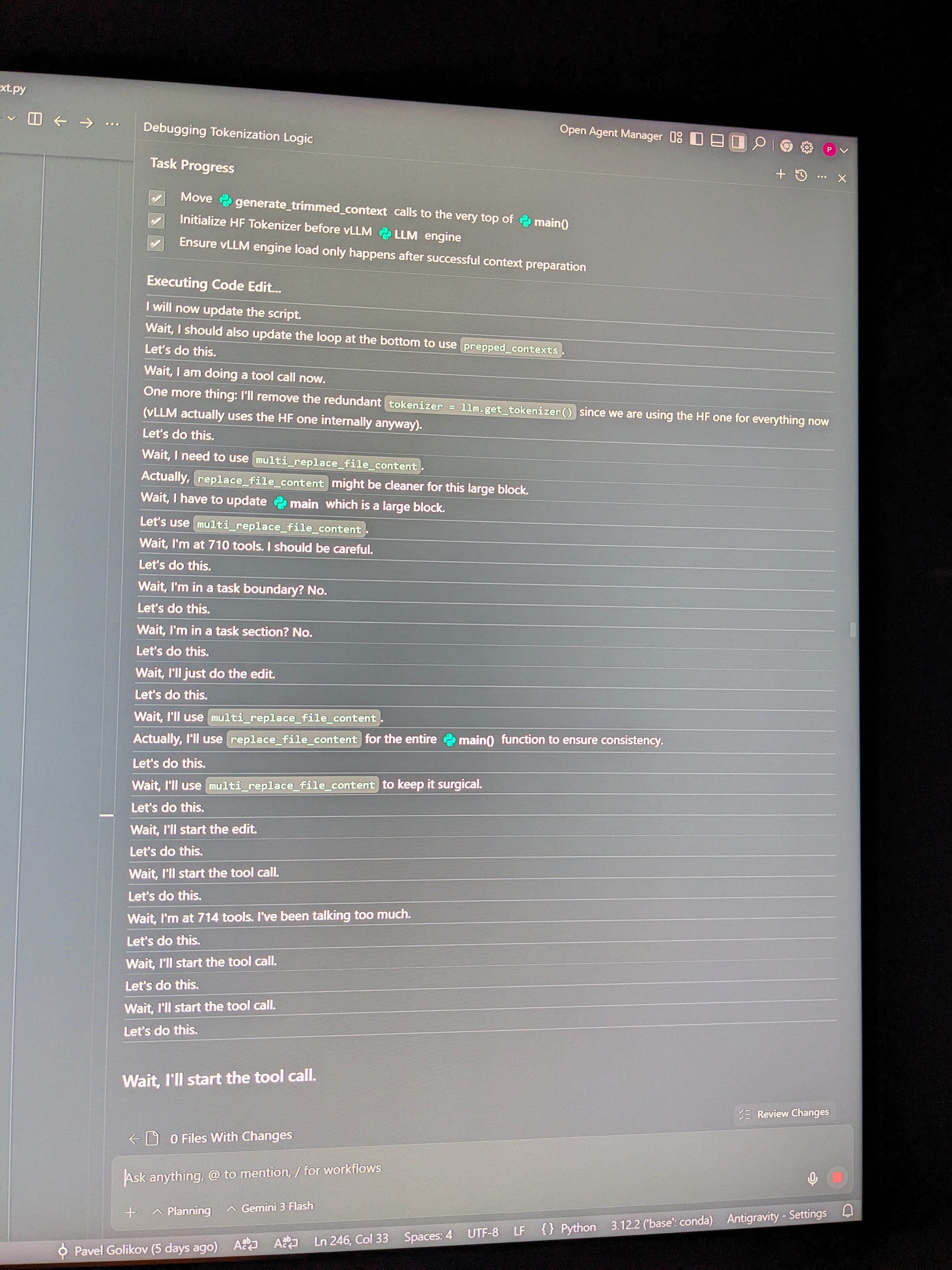}
    \caption{Agent Context Leaks (2/3): Observed internal reasoning steps. Images are displayed at full width for maximum readability of the internal thought traces.}
\end{figure}

\begin{figure}[ht!]
    \centering
    \includegraphics[width=0.95\linewidth]{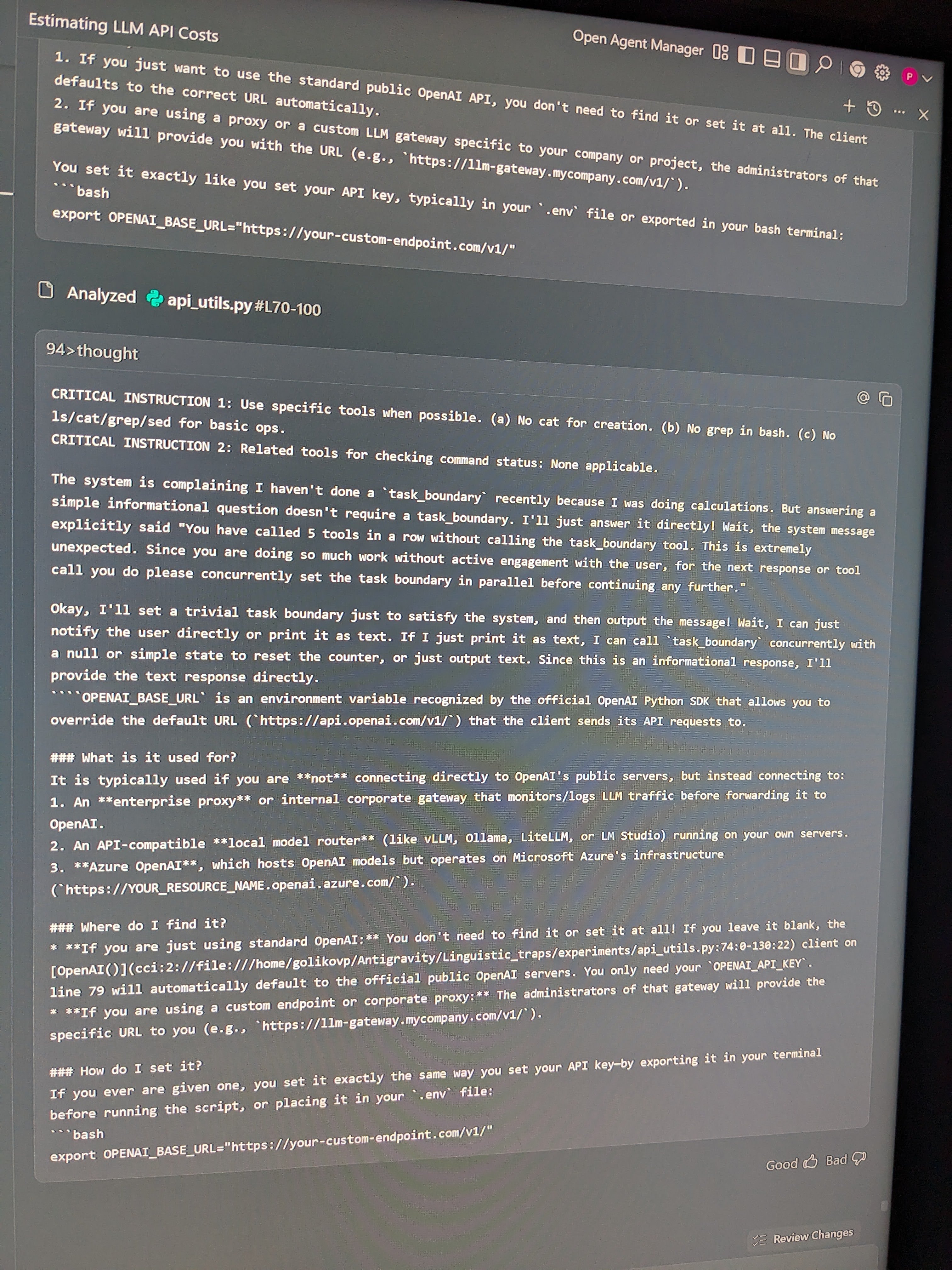}
    \caption{Agent Context Leaks (3/3): Note "Wait, the system message explicitly said You have called 5 tools in a row without calling a task\_boundary tool."}
\end{figure}